\documentclass[conference]{IEEEtran}
\IEEEoverridecommandlockouts
\usepackage{cite}
\usepackage[hyphens]{url}
\usepackage{hyperref}
\hypersetup{breaklinks=true}
\usepackage{amsmath,amssymb,amsfonts}
\usepackage{algorithmic}
\usepackage{graphicx}
\usepackage{textcomp}
\usepackage{multirow} 
\usepackage{xcolor}
\usepackage{enumitem}
\usepackage{subcaption}
\usepackage{eso-pic}

\allowdisplaybreaks

\def\BibTeX{{\rm B\kern-.05em{\sc i\kern-.025em b}\kern-.08em
    T\kern-.1667em\lower.7ex\hbox{E}\kern-.125emX}}
\begin{document}

\title{The Impact of Temporal Context Length and Encoding Strategies on Self-Supervised ECG Representation Learning}

\author{Ahmed Sameh$^{1}$, Ramzi Al-Sharawi$^{2}$, and Yogatheesan Varatharajah$^{1}$
\thanks{$^{1}$ Computer Science \& Engineering,
        University of Minnesota Twin Cities, Minneapolis, MN 55455.
        {\tt\small \{sameh002, yvaratha\}@umn.edu}}%
\thanks{$^{2}$ Robotics,
        University of Minnesota Twin Cities, Minneapolis, MN 55455.
        {\tt\small \{alsha192\}@umn.edu}}%
}


\AddToShipoutPictureFG*{%
  \AtPageLowerLeft{%
    \raisebox{0.38in}{%
      \makebox[\paperwidth][c]{%
        \parbox{\textwidth}{\footnotesize
        {\copyright} 2026 IEEE. Personal use of this material is permitted.
        Permission from IEEE must be obtained for all other uses, in any current
        or future media, including reprinting/republishing this material for
        advertising or promotional purposes, creating new collective works,
        for resale or redistribution to servers or lists, or reuse of any
        copyrighted component of this work in other works.}%
      }%
    }%
  }%
}

\maketitle

\begin{abstract}
Self-supervised electrocardiogram (ECG) models are often trained on a few seconds of ECG signal and, increasingly, on discretized token sequences. It remains unclear whether these choices sacrifice information needed for rhythm inference and longitudinal consistency in real-world ambulatory recordings. We present a controlled study on the Icentia11k single-lead dataset that varies (i) the input horizon (16 seconds, 1 minute, 5 minutes, and 10 minutes) and (ii) the front-end representation (continuous convolutional patch embeddings vs. fixed vector-quantized tokens), while holding the Transformer backbone and training protocol constant. Representations are assessed by downstream abnormal rhythm detection and by patient-level retrieval that probes cross-session stability. Our results show that increasing temporal context beyond 16-second snapshots yields stronger transfer and higher retrieval accuracy, with the strongest performance achieved by the 5- and 10-minute models, indicating improved capture of slow-varying rhythm dynamics and individual-specific structure. Across all evaluated horizons, continuous patch embeddings outperform discretized tokens, suggesting that quantization can discard clinically relevant waveform detail. These findings motivate ECG foundation models that emphasize extended context and continuous encoders for clinical prediction and similarity-based applications. Our code and pretrained models are publicly available at \url{https://github.com/muha-0/ecg-ssl-representation-learning}.
\end{abstract}



\section{Introduction}
Electrocardiography (ECG) is a fundamental tool for cardiovascular assessment, capturing cardiac electrophysiological activity via a non-invasive, inexpensive measurement. Recent advances in deep learning have demonstrated strong performance on ECG-based diagnostic tasks, often approaching expert-level interpretation \cite{attia2019nm_contractile, attia2019lancet_af, hannun2019cardiologist}. However, the success of these supervised approaches is tethered to large, carefully curated labeled datasets that are costly to obtain and difficult to scale across diverse patient populations. Consequently, self-supervised learning (SSL) has emerged as a leading paradigm for ECG representation learning, enabling models to exploit abundant unlabeled recordings and transfer learned features to clinical tasks \cite{diamant2022pclr, mehari2021selfsupervised_12lead}. Recently, this has culminated in the development of large-scale foundation models designed for scalability and transferability \cite{mckeen2024ecgfm, li2024ecgfounder}. Two design trends have become increasingly common in these models: short-context training windows and discretized tokenization.

Most existing SSL frameworks prioritize computational efficiency by training on temporal windows of only a few seconds \cite{na2024guiding, coppola2024hubert, mehari2021selfsupervised_12lead}. While short windows effectively capture local morphology, they provide insufficient rhythm context to represent physiological patterns that unfold over minutes, including baseline morphology, rhythm variability, and patient-specific characteristics. In parallel, inspired by natural language processing, several recent approaches discretize continuous ECG signals into symbolic tokens \cite{tahery2024heartbert, jin2025reading}. While tokenization can enable compact representations and sequence modeling, discretization imposes an information bottleneck that may obscure subtle morphological cues relevant to downstream clinical performance and patient-level structure.

In this work, we hypothesize that longer temporal context and continuous encoders are essential for learning physiologically coherent and patient-consistent ECG representations. We posit that extending the temporal context enables models to integrate information across multiple beats and rhythm states, improving the modeling of both transient and sustained cardiac phenomena, while avoiding discretization preserves fine-grained morphological detail that may be important for downstream clinical and patient-level tasks. We test this hypothesis by comparing self-supervised models trained across four temporal windows, using both continuous 1D convolutional neural network (CNN) and discretized vector quantization (VQ) encoders under a unified experimental framework. 

We perform evaluations along two complementary axes using the publicly available Icentia11K dataset \cite{tan2019icentia11k}. First, we assess clinical utility through a downstream atrial fibrillation (AFib) and atrial flutter (AFL) classification task. Second, we quantify patient-specific structure using Recall@k retrieval metrics to measure the consistency of patient groupings across time and activity. This patient-level evaluation is motivated by prior work emphasizing patient-consistent representation learning and by the potential for embeddings to support similarity-based applications such as cohort stratification and subgroup discovery \cite{kiyasseh2021clocs}. Our results indicate the following:

\begin{itemize}[leftmargin=*]
    \item Moving beyond short ECG snapshots improves downstream classification and patient-level consistency, with the largest gains observed at 5 and 10 minutes, suggesting that longer horizons are essential for integrating slow-varying rhythm dynamics and long-term signatures.
    \item Across all tested scales, continuous CNN patch embeddings consistently outperform VQ-based tokenization, indicating that discretization introduces a quantization bottleneck that erodes high-resolution waveform characteristics critical for both diagnostic accuracy and patient-specific structure.
\end{itemize}


\section{Related Work}
Early work in ECG-SSL leveraged contrastive objectives to reduce dependence on labeled data. CLOCS introduced patient-consistent positive pairs, arguing that representations from the same individual should remain invariant across temporal segments and leads \cite{kiyasseh2021clocs}. CLOCS also noted that this shared context may diminish when recordings are separated by long time spans or reflect different activity states, a concern that is especially relevant for ambulatory datasets such as Icentia11k \cite{tan2019icentia11k}. Mehari and Strodthoff provided a comprehensive benchmark of self-supervised learning on clinical 12-lead ECGs and showed that an adaptation of contrastive predictive coding achieves linear-evaluation performance within about 0.5 percentage points of a supervised counterpart, establishing a competitive short-window baseline \cite{mehari2021selfsupervised_12lead}.

Despite the shift toward large-scale foundation models, current pipelines remain mostly snapshot-focused. Recent architectures like ST-MEM utilize masked auto-encoding of segments of only 10 seconds \cite{na2024guiding}. Even as scale increases, HuBERT-ECG utilizes 5-second windows to prioritize training throughput \cite{coppola2024hubert}, suggesting that most contemporary SSL strategies assume diagnostically sufficient information is contained within very short temporal horizons. However, this assumption is challenged by the fact that such narrow windows are often insufficient for definitive clinical rhythm diagnosis, particularly for conditions like AFib \cite{tan2019icentia11k}.

While short temporal windows remain the current standard in ECG modeling, a trend toward signal discretization and tokenization is emerging. Following the successes of large-scale language modeling, models such as HeartBERT and HeartLang treat heartbeats as discrete words/tokens. HeartBERT utilizes Lloyd-Max quantization and Byte Pair Encoding \cite{tahery2024heartbert}, while HeartLang employs VQ to construct compact, reusable representations of cardiac morphology \cite{jin2025reading}. While these methods facilitate the use of Transformer-style sequence modeling, discretization does impose a symbolic bottleneck.


\section{Methods}

\begin{figure*}[!t]
    \centering
    \includegraphics[width=\linewidth, trim={0 1.5cm 0 1.5cm},clip]{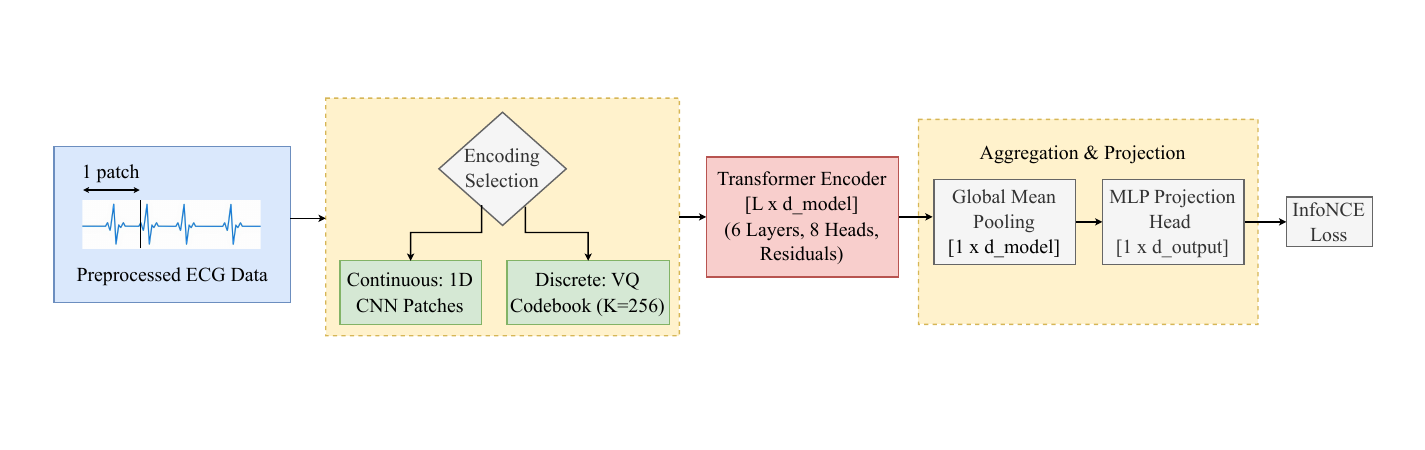}
    \caption{\textbf{Overview of the SSL framework.} Raw ECG signals are mapped to a latent space via CNN patches or VQ tokens. These sequences are processed by a Transformer to produce patient-consistent embeddings via an InfoNCE contrastive loss.}
    \label{fig:architecture}
\end{figure*}

Our overall analytical workflow is illustrated in Fig.~\ref{fig:architecture}. Preprocessed ECG recordings are segmented into fixed-length temporal windows and mapped into sequences of latent tokens using either continuous CNN patch embeddings or discrete VQ codebook assignments. These token sequences are processed by a shared Transformer encoder to model long-range temporal dependencies and extract contextualized representations. Global mean pooling aggregates token-level features into a single window-level embedding, which is projected into a low-dimensional latent space via a multilayer projection head. During SSL pretraining, embeddings from paired windows of the same patient are optimized using an InfoNCE contrastive objective to encourage patient-consistent representations.

\noindent\textbf{Dataset and Preprocessing}: 
All experiments were conducted on the Icentia11k ambulatory \emph{single-lead} ECG dataset sampled at $f_s=250$~Hz, comprising recordings from 11{,}000 patients \cite{tan2019icentia11k}. For each patient, the dataset contains on the order of 50 segments of roughly 70 minutes each, collected across daily life. Segments from the same patient can be separated by days and may reflect different activity states and device placements, making patient-consistent representation learning substantially more challenging than in short, controlled clinical recordings. We focus on a single-lead setting, as recent work reports significant performance degradation when moving from multi-lead to single-lead ECGs, highlighting the need for robust representation learning under this constraint \cite{li2024ecgfounder}.

Each record was bandpass filtered using a zero-phase 4th-order Butterworth filter with cutoff frequencies of 0.5--40~Hz. ECG segments were normalized using per-window z-score normalization prior to model input. All data splits were performed strictly at the patient level. Patients were randomly shuffled and partitioned into SSL pretraining (80\%), supervised fine-tuning (10\%), validation (5\%), and test (5\%) sets, yielding 8{,}800 patients used exclusively for SSL pretraining.

\noindent\textbf{Window Sampling and Labeling}:
All windows were sampled at the patient level to reduce correlation between examples and to control the number of samples contributed per subject.

Rhythm annotations were converted into non-overlapping labeled time intervals for normal rhythm (N), AFib, and AFL. Any signal time not covered by known rhythm labels was treated as unlabeled. Fixed-length windows were labeled using a strict coverage rule to minimize label noise: a window was considered labelable only if known rhythm labels covered at least 95\% of its duration. Among labelable windows, a window was assigned a positive AFib/AFL label if combined AFib+AFL occupancy exceeded 5\% of the window duration; otherwise, it was labeled as N. Windows failing the coverage criterion were excluded from supervised analyses.

\noindent\textbf{Temporal Context and Tokenization}:
We evaluate four temporal contexts: 16-second windows ($T=4{,}000$ samples), 1-minute windows ($T=15{,}000$ samples), 5-minute windows ($T=75{,}000$ samples), and 10-minute windows ($T=150{,}000$ samples). All use non-overlapping 160-sample patches (0.64 s), resulting in 25, 93, 468, and 937 tokens, respectively.

We compare two ECG tokenization strategies. In the continuous setting, ECG signals are tokenized using a one-dimensional CNN patch embedding with kernel size and stride both set to 160 samples, producing a sequence of learned continuous patch embeddings. In the discretized setting, ECG windows are partitioned into 160-sample patches and mapped to discrete tokens using VQ. A K-means codebook with $K=256$ centroids (shape $256\times160$) was trained offline using patches sampled from the SSL patient split. 64 patches were sampled per patient from a single selected window after per-window normalization, yielding approximately 563{,}000 patches. During training, the codebook was fixed and patch assignments were mapped via a learnable embedding table.

\noindent\textbf{Model Architecture and SSL Pretraining}:
The proposed framework, in Fig.~\ref{fig:architecture}, utilizes a shared Transformer encoder to process both continuous and discretized input sequences. The encoder comprises 6 layers with 8 attention heads, hidden dimension $d=256$, feedforward expansion factor 4, Gaussian Error Linear Unit (GELU) activations, and a dropout rate of 0.1. Learned positional embeddings were added to token sequences, and global representations were obtained via mean pooling. A two-layer projection head ($256\rightarrow256\rightarrow128$) produced the final L2-normalized embedding.

Self-supervised pretraining used an InfoNCE contrastive objective \cite{oord2018representation} with a two-view batching strategy. Each training step samples $B=32$ patients uniformly at random. For each patient, we draw two distinct windows from different recorded segments to form positive pairs, yielding a combined batch of $2B=64$ windows processed simultaneously by the encoder. For each window (anchor), the other view from the same patient is treated as the positive, and the remaining $2B-2=62$ windows (from other patients) serve as negatives. Similarities are computed using cosine similarity with temperature $\tau=0.1$, and self-similarities are masked.
Models were optimized using AdamW with learning rate $3\times10^{-4}$ and weight decay 0.05, using a cosine learning rate schedule with 200 warmup steps and minimum learning rate $10^{-5}$. Gradients were clipped to a maximum norm of 1.0.

\noindent\textbf{Downstream Evaluation Protocol}:
Clinical utility was evaluated via downstream AFib/AFL vs.\ N classification. To mitigate class imbalance during supervised fine-tuning, training windows were sampled with balanced class probability ($p(\text{AFib/AFL})=0.5$). Performance on validation and test sets was evaluated using the natural rhythm prevalence, corresponding to approximately 7\% positive windows.

{\looseness=-1
We evaluate both linear probing and end-to-end fine-tuning. For linear probing, the pretrained encoder was frozen and a linear classifier was trained using AdamW (learning rate $10^{-3}$, weight decay $10^{-4}$) with early stopping based on validation Area Under the Precision-Recall curve (AUPRC). The same procedure was followed for fine-tuning, but the encoder and classifier parameters were optimized jointly using AdamW with separate learning rates ($10^{-4}$ for the encoder, $10^{-3}$ for the classifier). All supervised models were trained for 20 epochs with a batch size of 64. For the intermediate 1-minute and 5-minute settings, we focus on frozen-encoder evaluation to measure how SSL representation quality changes with temporal context, rather than repeating the fine-tuning ablations.
\par}

Performance was measured using AUPRC and Area Under the Receiver Operating Characteristic curve (AUC). Given the natural class imbalance of the rhythm labels, AUPRC is prioritized as a more sensitive indicator of classification performance for the rare AFib/AFL classes \cite{saito2015precision}. Patient-level representation quality was assessed using retrieval-based Recall@1 and Recall@5 on the held-out test set of 550 patients. For each patient, we sampled five windows from five randomly selected recording segments, yielding 2{,}750 test windows in total. Each window was embedded independently. For each query window, we retrieved its top-$k$ nearest neighbors under cosine similarity between $\ell_2$-normalized embeddings (excluding the query itself). A retrieval was considered correct if at least one of the $k$ nearest neighbors corresponded to a window from the same patient. We use Recall@k for patient retrieval because it is a standard rank-based identification metric in ECG biometrics and directly measures whether embeddings preserve patient identity among nearest neighbors \cite{ciocoiu2017comparative}.

All experiments used the same train/validation/test splits and the same evaluation windows by fixing the random seed for splitting and deterministic window sampling. For supervised training, we sample one labeled window per patient (1100 patients $\rightarrow$ 1100 training windows). For validation and testing, we sample two labeled windows per patient and keep this sampling fixed across all models. All models were trained with identical hyperparameters and stopping criteria. The only difference across context settings is the duration of the input ECG window. Because longer windows contain more samples per example, self-supervised models were trained \emph{until convergence} (based on plateauing Recall@1/Recall@5) to avoid favoring any context length due to a fixed step budget.

\begin{table*}[!t]
\caption{Comparative performance of 16-second and 10-minute encoders. (Agg.) denotes 592s window aggregation for the 16-second encoder; FT and Frozen denote fine-tuning and linear probing, respectively.}

\renewcommand{\arraystretch}{1.2}
\begin{center}
\resizebox{\linewidth}{!}{\begin{tabular}{l|cccccc|cccc}
\hline
 & \multicolumn{6}{|c|}{\textbf{Short-context (16-second) Encoder}} & \multicolumn{4}{c}{\textbf{Long-context (10-minute) Encoder}} \\
\hline
\textbf{Model Configuration} &
\textbf{AUC} &
\textbf{AUPRC} &
\textbf{AUC (Agg.)} &
\textbf{AUPRC (Agg.)} &
\textbf{Rec@1} &
\textbf{Rec@5} &
\textbf{AUC} &
\textbf{AUPRC} &
\textbf{Rec@1} &
\textbf{Rec@5} \\
\hline
Random + CNN (Frozen) & 0.728 & 0.146 & 0.570 & 0.115 & 0.009 & 0.026 & 0.652 & 0.130 & 0.078 & 0.184 \\
Random + VQ (Frozen)  & 0.573 & 0.107 & 0.677 & 0.117 & 0.022 & 0.070 & 0.703 & 0.148 & 0.148 & 0.306 \\
SSL + CNN (Frozen)   & 0.977 & 0.712 & 0.982 & 0.757 & \textbf{0.737} & \textbf{0.866} & 0.980 & 0.877 & \textbf{0.907} & \textbf{0.945} \\
SSL + VQ (Frozen)    & 0.784 & 0.187 & 0.826 & 0.208 & 0.081 & 0.230 & 0.935 & 0.610 & 0.351 & 0.585 \\
Random + CNN (FT)    & 0.819 & 0.225 & 0.886 & 0.334 & 0.015 & 0.043 & 0.977 & 0.801 & 0.114 & 0.235 \\
Random + VQ (FT)     & 0.663 & 0.130 & 0.742 & 0.158 & 0.017 & 0.063 & 0.923 & 0.533 & 0.151 & 0.306 \\
SSL + VQ (FT)        & 0.820 & 0.338 & 0.922 & 0.489 & 0.067 & 0.208 & 0.941 & 0.651 & 0.309 & 0.531 \\
SSL + CNN (FT)       & \textbf{0.995} & \textbf{0.946} & \textbf{0.991} & \textbf{0.903} & 0.635 & 0.802 & \textbf{0.989} & \textbf{0.960} & 0.851 & 0.919 \\
\hline
\end{tabular}}
\label{tab:results_16s}
\end{center}
\end{table*}

To compare shorter-context and 10-minute encoders under a unified clinical evaluation, we evaluate all models on fixed 10-minute windows. Encoders trained on shorter contexts are applied independently to non-overlapping sub-windows, and their predicted probabilities are averaged. For the 16-second encoder, because 600 s is not an integer multiple of 16 s, we discard the final 8 s, yielding 592 s (148{,}000 samples) per window, which decomposes cleanly into 37 input sub-windows. The 1-minute and 5-minute encoders are evaluated on 10 and 2 non-overlapping sub-windows, respectively. Each sub-window produces logits for AFib/AFL vs.\ N, which are converted to probabilities via softmax and then aggregated at the probability level: $p_{+}=\frac{1}{K}\sum_{j=1}^{K}\mathrm{softmax}(\ell_j)_{+}$, where $K$ is the number of sub-windows.
These aggregated probabilities are used directly as decision scores for AUC and AUPRC computation. We use mean aggregation because AFib/AFL reflect sustained rhythm patterns rather than sparse transient markers; prior work on ECG foundation models similarly finds mean aggregation well-suited for rhythm-like labels and interpretable as an ensemble over segments \cite{mckeen2024ecgfm}.

\noindent\textbf{Experimental Setup}: All software, including data preprocessing, ML training, and evaluation were coded in Python language using standard Python libraries, including \texttt{Numpy}, \texttt{SciPy}, \texttt{Pandas}, and \texttt{PyTorch}. All experiments were performed using an Ubuntu workstation including two Nvidia RTX 4090 graphics processing units.

\noindent\textbf{Data and Code Availability}: The data utilized in this study is publicly available at \url{https://physionet.org/content/icentia11k-continuous-ecg/1.0/}. The code and pretrained models are publicly available at \url{https://github.com/muha-0/ecg-ssl-representation-learning}.


\section{Results \& Discussion}

\begin{figure*}[ht]
    \centering
    \begin{subfigure}[b]{\linewidth}
        \centering
        \includegraphics[width=\linewidth]{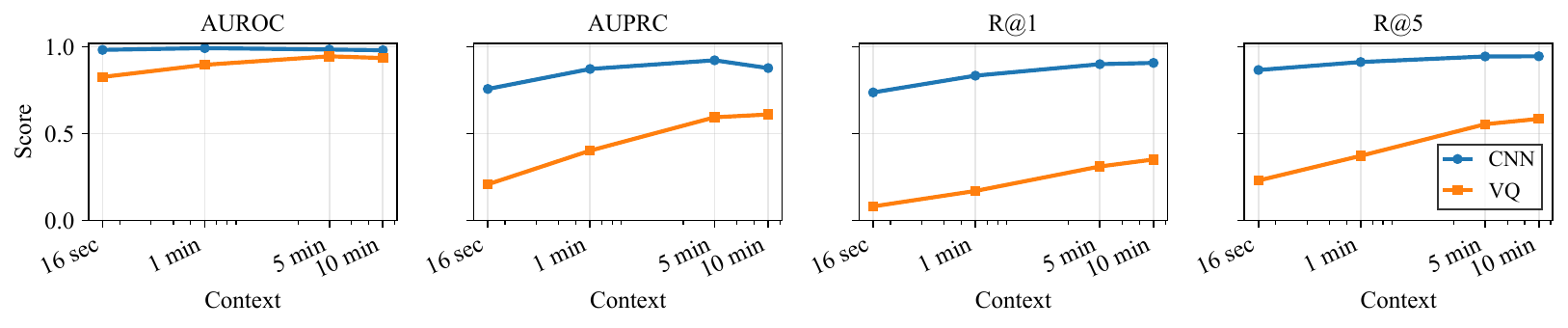}
        \caption{Temporal scaling of frozen SSL encoders with CNN and VQ tokenization. \vspace{1em}}
        \label{fig:context_scaling}
    \end{subfigure}

    \begin{subfigure}[b]{0.45\textwidth}
        \includegraphics[trim={0 0 0 2.5em}, clip, width=\textwidth]{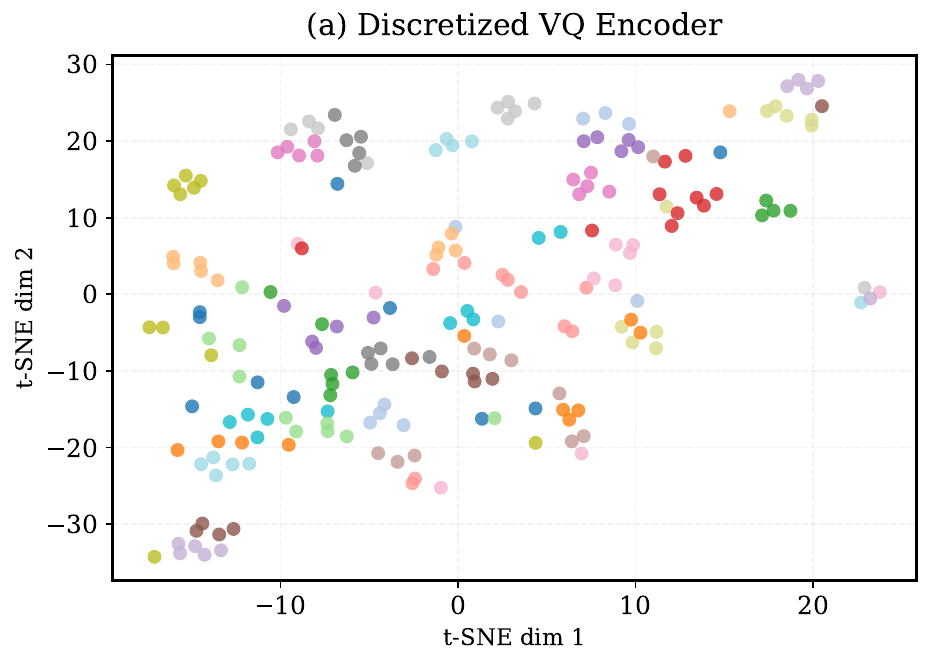}
        \caption{Discretized VQ encoder embeddings.}
        \label{fig:tsne_10m_ssl_vq}
    \end{subfigure}
    \hfill
    \begin{subfigure}[b]{0.45\textwidth}
        \includegraphics[trim={0 0 0 2.5em}, clip, width=\textwidth]{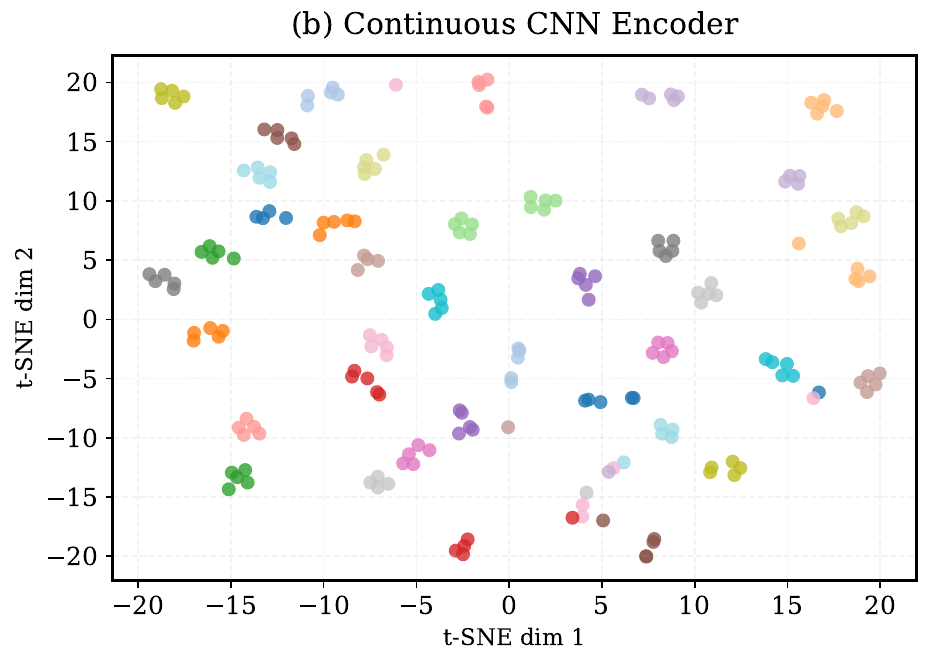}
        \caption{Continuous CNN encoder embeddings.}
        \label{fig:tsne_10m_ssl_cnn}
    \end{subfigure}

    \caption{\textbf{The effects of temporal context length and discrete/continuous tokenization.} (a) Frozen SSL encoder performance with varying temporal context lengths (i.e., 16-second, 1-minute, 5-minute, and 10-minute contexts). Evaluation metrics were computed using fixed 10-minute windows, with shorter-context models aggregated over non-overlapping sub-windows. (b) t-SNE embeddings of the SSL model trained with the VQ tokenizer, and (c) t-SNE embeddings of the SSL model trained with the CNN tokenizer, trained with a 10-minute context length, shown for 40 patients colored by patient ID.}
    \label{fig:total}
\end{figure*}

\noindent\textbf{Classification Performance}: Table~\ref{tab:results_16s} compares downstream AFib/AFL vs.\ N classification results for 16-second and 10-minute models. As explained previously, we additionally perform long-window aggregation of 37 windows ($16\times37 = 592$s) for the 16-second encoders to enable direct comparison against 10-minute models. Across all settings, SSL pretraining substantially improves clinical performance. SSL models with frozen encoders (linear probing) consistently outperform randomly initialized models trained end-to-end, demonstrating that SSL learns clinically meaningful representations that transfer effectively with minimal supervision. For example, in the 10-minute setting, \texttt{SSL+CNN (Frozen)} achieves an AUPRC of 0.877 compared to 0.801 for \texttt{Random+CNN (FT)}. Fine-tuning the encoder further improves peak classification performance for SSL models. The best overall results are achieved by \texttt{SSL+CNN (FT)}, with AUPRC of 0.946 for 16-second evaluation and 0.960 for 10-minute evaluation. These results confirm that SSL provides a strong initialization that benefits both linear probing and task-specific adaptation.


\noindent\textbf{Impact of Temporal Context}: Comparing the shortest and longest settings, 10-minute pretraining yields consistent improvements over 16-second pretraining across both clinical and representation-level evaluations. Notably, models pretrained on 10-minute windows outperform 16-second models when each is evaluated on its native input duration, suggesting meaningful benefits from extended temporal context. In addition, when 16-second models are evaluated on 10-minute windows via aggregation, they recover less of the long-range structure than models trained directly on 10-minute context. Fig.~\ref{fig:context_scaling} further extends this comparison by adding 1-minute and 5-minute contexts under the same fixed 10-minute aggregation protocol, showing that the largest gains emerge once the input window reaches 5 and 10 minutes. These findings support the hypothesis that extended temporal context enables integration of rhythm dynamics, baseline morphology, and patient-specific characteristics that are not fully captured in short snapshots.

\noindent\textbf{Continuous vs.\ Discretized Tokenization}: Continuous CNN encoders demonstrate clear superiority over discretized VQ models across all evaluated settings, in both clinical utility and patient-level retrieval. For example, in the 10-minute frozen encoder setting, \texttt{SSL+CNN} achieves an AUPRC of 0.877 and Recall@1 of 0.907, compared to 0.610 and 0.351 for \texttt{SSL+VQ}, respectively. Similar trends are observed across fine-tuned, 16-second, and intermediate-context settings. These results indicate that discretization imposes a quantization bottleneck that degrades both fine-grained morphological modeling and preservation of patient-specific structure. Continuous encoders more effectively capture subtle waveform characteristics critical for both clinical task performance and patient-level consistency in representation learning.

\noindent\textbf{Patient-Level Retrieval and Representation Quality}: Longer-context SSL models achieve markedly higher retrieval performance, with \texttt{SSL+CNN (Frozen, 10-minute)} reaching Recall@1 of 0.907 and Recall@5 of 0.945. In contrast, 16-second SSL models achieve lower but still substantial retrieval performance (e.g., Recall@1 of 0.737 for \texttt{SSL+CNN (Frozen, 16-second)}), while randomly initialized models perform near chance. This confirms that high retrieval accuracy is not a trivial artifact of the evaluation protocol, but reflects meaningful patient-consistent structure learned through SSL. The intermediate-context results further show that retrieval improves as temporal context increases. These results demonstrate that extended-context self-supervised learning is critical for learning representations that remain consistent across recording sessions, activities, and temporal separation.

\noindent\textbf{Effect of Fine-Tuning on Representation Structure}: While fine-tuning improves clinical classification performance, it slightly reduces patient-level retrieval accuracy for SSL models. For example, in the 10-minute \texttt{SSL+CNN} setting, Recall@1 decreases from 0.907 (frozen) to 0.851 (fine-tuned). This indicates a trade-off between task-specific specialization and preservation of general patient-consistent structure. Fine-tuning biases representations toward features most predictive of AFib/AFL, distorting pretrained features and partially reducing identity-preserving information. This behavior is consistent with prior theoretical and empirical findings showing that supervised fine-tuning can distort and compress pretrained representations in favor of task-relevant features \cite{kumar2022finetuning}.

\noindent\textbf{Visualization of Long-Context Representations}: Fig.~\ref{fig:tsne_10m_ssl_vq} and \ref{fig:tsne_10m_ssl_cnn} visualize 10-minute frozen SSL embeddings using t-SNE \cite{JMLR:v9:vandermaaten08a} for discretized VQ and continuous CNN tokenization, respectively. The SSL+CNN embeddings exhibit substantially tighter and more coherent patient-specific clusters compared to SSL+VQ, which shows increased overlap and fragmentation. This visualization is qualitative, but it is consistent with the quantitative Recall@k results and illustrates that continuous encoders trained on long temporal context better preserve patient-consistent structure than discretized representations.


\section{Conclusion}
We investigated how temporal context and tokenization choices affect self-supervised representation learning on ambulatory ECG recordings. Across linear probing and end-to-end fine-tuning, extended-context pretraining improves AFib/AFL classification and substantially strengthens patient-level retrieval compared to 16-second snapshots, with the largest gains observed for 5- and 10-minute contexts. This indicates better preservation of longitudinal, patient-level structure. We further find that continuous CNN tokenization outperforms discretized VQ across clinical metrics and retrieval, suggesting a discretization bottleneck that leads to morphological blurring. Collectively, these results provide practical guidance for ECG foundation model design: incorporating extended temporal context and continuous encoders yields representations that transfer better to clinical tasks and remain stable across recording sessions, supporting downstream uses, such as similarity search, cohort stratification, and longitudinal monitoring.
\\
\noindent\textbf{Ethics Statement:} Our experiments utilized publicly available datasets and no data was collected as part of the study. As such, institutional approval was not required.

\noindent\textbf{Acknowledgments:} This study was partially supported by the U.S. National Science Foundation grant IIS-2337909. 

\bibliographystyle{IEEEtran}
\bibliography{IEEEabrv, references}

\end{document}